\documentclass[10pt,twocolumn,letterpaper]{article}
\usepackage[pagenumbers]{cvpr}
\usepackage{newtxmath}
\usepackage{wrapfig}
\usepackage{newtxtext,newtxmath}
\usepackage{tgheros}
\usepackage{tgcursor}
\usepackage{colortbl}
\usepackage{multirow} 
\usepackage{bm}
\usepackage[most]{tcolorbox}
\newtcolorbox{questionbox}{enhanced,
  colback=white, colframe=black!80, boxrule=1.2pt, arc=2.5mm,
  drop shadow={black!55, opacity=1},
  halign=center, fontupper=\itshape,
  left=6pt, right=6pt, top=4pt, bottom=4pt,
  before skip=6pt, after skip=6pt}

\usepackage{xcolor}
\usepackage{ulem}

\usepackage{booktabs}

\usepackage{booktabs}
\usepackage{amsmath,amssymb}
\usepackage{comment}
\usepackage{amsthm}
\newtheorem{proposition}{Proposition}
\newtheorem{corollary}{Corollary}
\usepackage{multirow}
\definecolor{cvprblue}{rgb}{0.21,0.49,0.74}
\usepackage[pagebackref,breaklinks,colorlinks,allcolors=cvprblue]{hyperref}
\def\confName{CVPR}
\def\confYear{2027}

\newcommand{\CubeRelErrRel}{$-$23}

\newcommand{\CubeRelQselCI}{+0.02\,[$-$0.47, +0.50]}

\newcommand{\CubeRelBetter}{13}
\newcommand{\CubeRelWorse}{16}

\newcommand{\NSelSig}{7}

\newcommand{\ErrRelMax}{23}

\newcommand{\ThSharedMin}{35}
\newcommand{\ThSharedMax}{83}

\newcommand{\TPushTPGap}{26}

\newcommand{\TPushTDinoPGap}{57}

\newcommand{\TTwoRoomBGap}{7}

\newcommand{\TWallBGap}{17}

\newcommand{\HNineMean}{+0.5}

\newcommand{\TPushTPPredFailStart}{61}
\newcommand{\TPushTPPredFailTrained}{46}

\newcommand{\TPushTDinoPPredFailStart}{49}
\newcommand{\TPushTDinoPPredFailTrained}{20}

\usepackage{tabularx}

\title{When Do Better Visual World Models Make Better Plans?\\ Decision-Aligned Selection for Latent Planning}
\author{%
  \textbf{Chan Aristella Lu}\textsuperscript{1} \quad
  \textbf{Arya Fayyazi}\textsuperscript{2} \quad
  \textbf{Junil Yeo}\textsuperscript{1} \quad
  \textbf{Jin Lu}\textsuperscript{1}\\[2pt]
  \textbf{Kyu Hyung Lee}\textsuperscript{1} \quad
  \textbf{Mehdi Kamal}\textsuperscript{2} \quad
  \textbf{Massoud Pedram}\textsuperscript{2}\\[4pt]
  \textsuperscript{1}University of Georgia \quad
  \textsuperscript{2}University of Southern California
}

\begin{document}
\twocolumn[{%
\renewcommand\twocolumn[1][]{#1}%
\maketitle
\begin{center}
  \centering
  \includegraphics[width=\textwidth,height=5in,keepaspectratio]{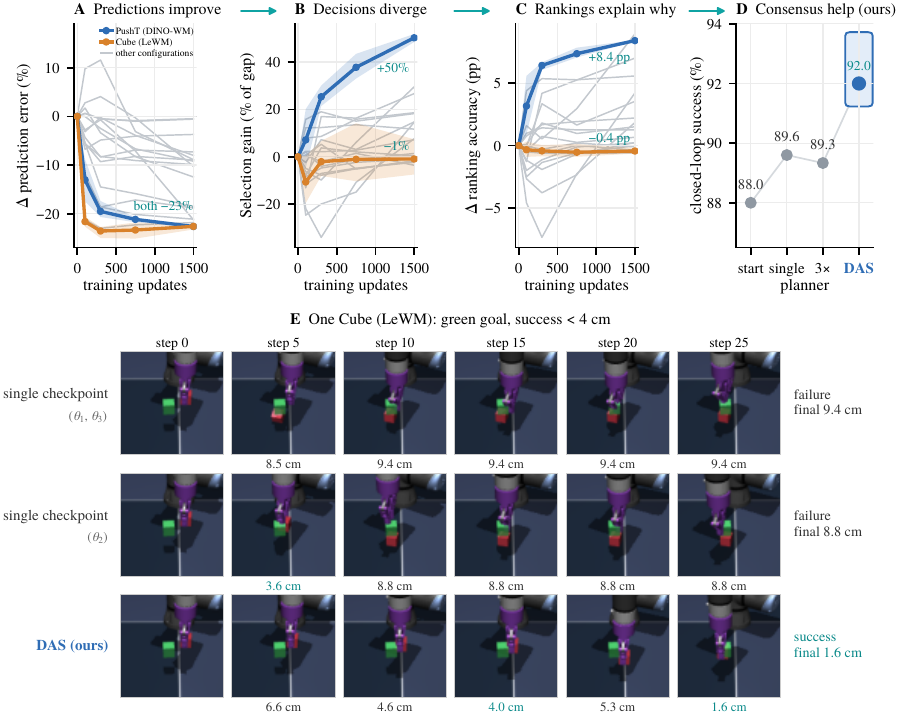}
  \vspace{-6pt}
  \captionof{figure}{\textbf{Equal prediction gains, unequal decisions.}
(A--C)~Two configurations whose prediction error falls by the same amount on the development pools: PushT with DINO-WM (blue) and Cube with LeWM (orange),
mean and range of three seeds; grey: all other configurations (mean). All checkpoints score the same planner candidates, each of which is executed
in the simulator, so differences come from the models alone.
(A)~Continued training lowers the prediction error by 23\% in both.
(B)~Selection improves on PushT, closing 50\% of the gap to selection with the true outcomes, but not on Cube.
(C)~Ranking accuracy, how often the model orders two candidates correctly, explains why: on Cube, training removed errors that do not
change which candidate looks best.
(D)~Decision-aligned selection (DAS) scores candidates by the consensus of three checkpoints $\theta_1,\theta_2,\theta_3$ trained with
different seeds, with a penalty on their disagreement. In closed loop on PushT (LeWM), success rises from 88.0\% before training and
89.6\% for a single checkpoint to 92.0\%; a single checkpoint given three times the samples reaches 89.3\%.
(E)~On a Cube (LeWM) state, every single checkpoint selects a failing plan ($\theta_1$ and $\theta_3$ pick the same one);
DAS selects one that ends 1.6\,cm from the goal.
}
  \label{fig:teaser}
\end{center}
\vspace{4pt}
}]
\begin{abstract}
Visual world models enable goal-conditioned planning by predicting the outcomes of candidate actions in latent space.
However, planning depends on how well a model ranks candidates, whereas training optimizes how accurately it predicts each one.
We identify the Decision-Relevant Residual (DRR), the part of the cost error that can reorder candidates, and show that the more training reduces the DRR, the more it improves selection (Spearman $\rho=0.82$, vs.\ $0.61$ for prediction error).
Building on this insight, we introduce Decision-Aligned Selection (DAS), a plug-in selection rule that scores candidates by the standardized consensus of several independently trained checkpoints, which averages out their checkpoint-specific residuals.
Across two model families (LeWorldModel and DINO-WM) and five tasks, DAS selects better actions than single checkpoints in 12 of 18 configurations and outperforms training one model three times longer in 14 of 18.
In closed loop, DAS improves success over single checkpoints in 6 of 9 pre-registered LeWM comparisons.
\end{abstract}
\vspace{-20pt}
\section{Introduction}
\label{sec:intro}
Recent visual world models for planning increasingly follow the joint-embedding predictive architecture (JEPA)~\cite{lecun2022path,assran2023ijepa}: an encoder maps images to latent states, and a predictor rolls these states forward under candidate actions without reconstructing pixels~\cite{maes2026lewm,assran2025vjepa2,sobal2025pldm,zhou2024dinowm}.
Given a goal image, a planner samples action sequences and executes the one whose predicted outcome lies closest to the goal in latent space.
World models are thus trained and compared by prediction error, on the assumption that better predictions yield better plans.

However, lower prediction error does not guarantee better selection.
In Fig.~\ref{fig:teaser}A--C, continued training lowers the prediction error of two configurations by the same amount, yet selection improves only on PushT; on Cube, the trained model ranks the candidates no better than before.
Prior studies report the same gap: prediction error tracks planning success poorly~\cite{maes2026swm,terver2025jepawm,min2026lowerror}.

Existing work has approached this gap in two ways.
One tries to close it by retraining the model to be decision-aware, through new objectives~\cite{min2026lowerror,wang2026dalewm,chahe2026mcr} or planner-generated data~\cite{agrawal2025traintest}, but its benefit to selection in closed loop remains unconfirmed~\cite{min2026lowerror,agrawal2025traintest}.
The other measures it by closed-loop success~\cite{zhang2025worldinworld}, but a different model also searches different candidates and reaches different states, so a higher success rate cannot tell whether the model selects better or merely searches differently.
Neither route isolates which prediction gains improve selection.
We therefore ask: \begin{questionbox}
When do better predictions make better plans?
\end{questionbox}

To answer it, we compare models on \emph{fixed pools}: we freeze the planner's final candidates and execute each one in the simulator, so every model is judged on the same outcomes.
Our key observation is that the planner only compares candidates.
An error that raises or lowers the predicted costs of all candidates together leaves its choice unchanged; only the remaining error can reorder them.
We call this part the \emph{Decision-Relevant Residual} (DRR, \cref{sec:theory}).
In practice, error shared across candidates dominates prediction error~\cite{min2026lowerror}, so a model can predict much better without reducing the DRR at all. Building on this view, we propose \emph{Decision-Aligned Selection} (DAS, \cref{sec:method}).
The planner scores each candidate by the consensus of several checkpoints from a multi-seed training run.
This averages out the errors specific to each checkpoint and steers selection toward better candidates (Fig.~\ref{fig:teaser}E).
Our contributions are as follows:
\begin{itemize}[leftmargin=*,itemsep=1pt,topsep=2pt]
\item We identify the \textbf{Decision-Relevant Residual} (DRR) as the part of prediction error that governs the planner's choice. Across 18 model--task configurations, its reduction predicts which training gains improve selection more reliably than prediction error does (Spearman $\rho{=}0.82$ vs.\ $0.61$) (\cref{sec:diagnosis}).
\item We propose \textbf{Decision-Aligned Selection} (DAS), a plug-in selection rule that combines the checkpoints of a multi-seed training run at planning time and steers the planner toward better candidates (\cref{sec:method}).
\item We demonstrate strong gains across two model families and five tasks: DAS outperforms training one model three times longer in 14 of 18 configurations and lifts closed-loop success on PushT from 89.6\% to 92.0\% (\cref{sec:experiments}).
\end{itemize}

\begin{figure*}[h]
  \centering \includegraphics[width=\textwidth]{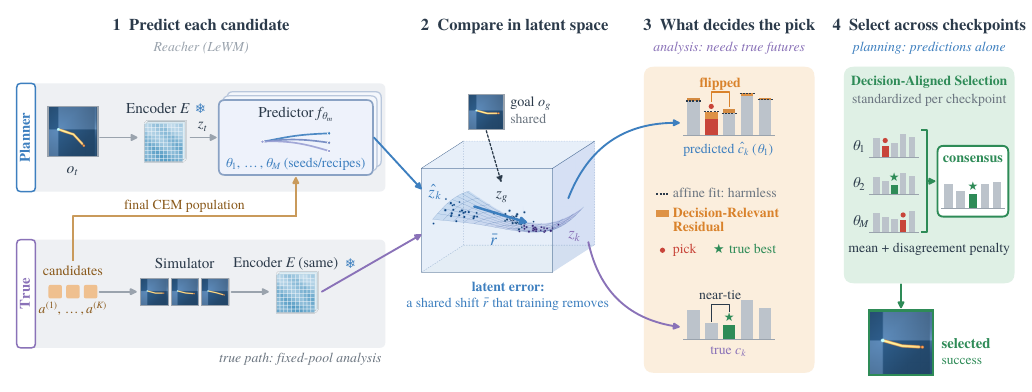}
  \caption{\textbf{Overview.}
  (1)~Each checkpoint $\theta_m$ predicts the latent outcome $\hat z_k$ of every candidate. On fixed pools, each candidate is also executed and encoded, giving $z_k$ (\cref{sec:prelim}).
  (2)~The latent error is dominated by a shared shift $\bar r$.
  (3)~The best affine fit (dashed) of the costs cannot change the pick. The Decision-Relevant Residual (orange) flips a near-tie, so the argmin misses the true best (\cref{sec:theory}).
  (4)~Decision-Aligned Selection combines the standardized costs of $M$ checkpoints
 (\cref{sec:method}).}
  \label{fig:overview}
  \vspace{-6mm}
\end{figure*}

\vspace{-10pt}
\section{Related Work}
\label{sec:related}

\textbf{Visual world models for planning.}
For visual goal-conditioned planning, latent world models predict the outcomes of candidate action sequences and score them against a goal embedding.
This study focuses on JEPA-style models, which predict in the space of frozen pre-trained features~\cite{zhou2024dinowm} or of representations learned end to end~\cite{maes2026lewm,assran2025vjepa2,sobal2025pldm}.
Generative models such as Navigation World Models (NWM)~\cite{bar2025nwm} instead synthesize future latents with diffusion.
Planning typically uses sampling-based optimizers such as the cross-entropy method (CEM)~\cite{rubinstein1999cem,pinneri2020icem} or model predictive path integral control (MPPI)~\cite{williams2017mppi}.

\textbf{Prediction accuracy and planning quality.}
World-model prediction accuracy and planning performance often diverge, both in model-based reinforcement learning~\cite{lambert2020objective} and in visual world models~\cite{maes2026swm,terver2025jepawm,zhang2025worldinworld}.
Prior analyses compare candidate rankings with true outcomes~\cite{wang2026dalewm}, trace planning failures to the latent objective~\cite{singh2026objective}, or bound planning regret in terms of pretraining error~\cite{cui2026jepatheory}.
Closely related to our work, Min \etal~\cite{min2026lowerror} show that error shared across candidates dominates prediction error but has little effect on candidate comparisons.
We take this analysis from diagnosis to remedy: we isolate the DRR, the part of prediction error that can alter the planner's choice, show that its reduction predicts when training gains translate into better action selection, and use it to design DAS, which improves selection at planning time.
\vspace{-1.5pt}

\textbf{Decision-aware model learning.}
Decision-focused and value-aware learning train models for the downstream decision~\cite{donti2017task,mandi2023dfl,grimm2020value}.
For world models, this includes learning monotone planning costs~\cite{chahe2026mcr}, retraining on planner-generated data~\cite{agrawal2025traintest} and test-time adaptation~\cite{mete2026optta}.
These methods improve the model itself; DAS is complementary, improving how the planner selects among the model's predictions.

\textbf{Ensembles for robust planning.}
Taking the minimum of noisy estimates favors optimistic errors, the optimizer's curse~\cite{smith2006optimizer}, and planners likewise exploit the errors of imperfect world models~\cite{bhamidipaty2026exploitable}.
Ensembles account for uncertainty during planning~\cite{chua2018pets} or penalize disagreement in offline reinforcement learning~\cite{yu2020mopo}, and learned metrics re-rank the candidates of a fixed world model~\cite{li2026trm}.
DAS instead acts only in the planner's scoring step: it scores candidates by the consensus of the checkpoints' standardized costs, optionally penalizing their disagreement.
We find that its gain grows with the diversity of their decision-relevant errors.

\section{Prediction Error and Selection}
\label{sec:diagnosis}

\subsection{Evaluation Protocol}
\label{sec:prelim}

\paragraph{Fixed pools.}
The world model predicts the latent $\hat z_k$ reached $H{=}25$ steps after candidate $a_k$, and the planner executes $j=\arg\min_k \hat c_k$ with $\hat c_k=\|\hat z_k-z_g\|^2$ and goal latent $z_g$.
At each state, we freeze the final cross-entropy method (CEM) population, draw $K{=}128$ candidates and execute each one in the simulator, so all checkpoints share the same outcomes (Fig.~\ref{fig:overview} left).
Each candidate thus has a true cost $c_k=\|z_k-z_g\|^2$ computed from its encoded real outcome $z_k$ and a physical quality $q_k$ equal to its task error at step 25 (lower is better).
With $k^\star=\arg\min_k c_k$ and $m=\arg\min_k q_k$, the regret splits into
\begin{equation*}
\vspace{-1mm}
q_j-q_m=\underbrace{(q_j-q_{k^\star})}_{\text{selection}}+\underbrace{(q_{k^\star}-q_m)}_{\text{objective}} .
\vspace{-2.5mm}
\end{equation*}
Only the selection part depends on the world model, and it can be negative.

\textbf{Metrics and training.}
The \emph{\% of gap} is the reduction in the mean error $q_j$ of the selected candidate, as a share of the starting model's mean gap $q_j-q_{k^\star}$.
A pick is \emph{near-optimal} if its error is within a quarter of the success radius of $q_m$, the best in the pool.
Pairwise ranking accuracy compares $\hat c$ with $q$ over candidate pairs whose quality differs by more than this tolerance.
A wrong pick is a \emph{prediction failure} if selection with true futures is near-optimal, and an \emph{objective failure} otherwise.
From each released model, we continue training the dynamics for 1{,}500 updates with three seeds and three equal-budget recipes.
Recipe $B$ mixes executed candidates with dataset windows and combines distillation towards the starting model with regression to real futures.
Recipe $O$ keeps this loss but drops the executed candidates, and $P$ is plain continued training on dataset windows.
This section uses development pools; test pools are reserved for \cref{sec:experiments}.

\subsection{Motivating Evidence}
\label{sec:observation}
\begin{figure}[t]
  \centering
  \includegraphics[width=0.8\linewidth]{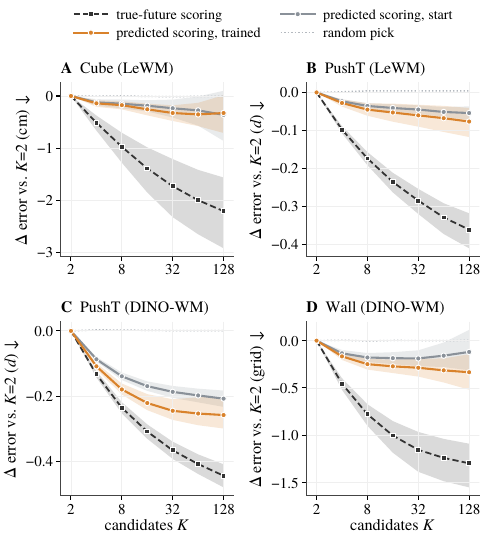}
  \caption{\textbf{More candidates help true futures far more than predictions} (development pools).
  True futures are scored with the same latent objective. Bands: 95\% CIs.}
  \label{fig:candcount}
  \vspace{-15pt}
\end{figure}

On the development pools, continued training lowers prediction error in all 18 configurations, by up to \ErrRelMax\%.
Yet it improves selection significantly in only \NSelSig.
On Cube, recipe $B$ changes the error by \CubeRelErrRel\%, but selected quality stays unchanged (95\% CI \CubeRelQselCI\,cm).
Training changes 80\% of the picks, but only \CubeRelBetter\% improve and \CubeRelWorse\% get worse; the rest are ties.
Across configurations, the change in ranking accuracy tracks the selection gain far better than error reduction does (Spearman $\rho{=}0.85$ vs.\ $0.46$).
For the starting models, prediction failures outnumber objective failures in every model--task pair ($28$--$94\%$ vs.\ $0$--$20\%$ of states).
Selection with predictions also saturates as candidates are added, while selection with true futures keeps improving (\cref{fig:candcount}).
These results point to how the model compares candidates, not how accurately it predicts each one.

\subsection{Decision-Relevant Residual}
\label{sec:theory}

Within a single state, we explain why lower prediction error does not necessarily change the pick and why selection saturates as candidates are added.
For a state with $K$ candidates, decompose the prediction residual as $r_k=\hat z_k-z_k=\bar r+\delta_k$, with $\bar r$ shared by all candidates and $\sum_k\delta_k=0$.

\begin{proposition}[Shared error is nearly rank-irrelevant; cf.~\cite{min2026lowerror}]
\label{prop:shared}
With $\bar{\hat z}$ the mean prediction and $\kappa$ a constant that is the same for every candidate,
\begin{equation*}
\hat c_k = c_k + 2\bar r^{\top}(\hat z_k-\bar{\hat z}) + 2\delta_k^{\top}(z_k-z_g) + \|\delta_k\|^2 + \kappa .
\end{equation*}
The shared residual changes the comparison of two candidates $k,l$ only through $2\bar r^{\top}(\hat z_k-\hat z_l)$, which is at most $2\|\bar r\|\max_{k,l}\|\hat z_k-\hat z_l\|$, whereas the squared prediction error counts it in full: $\frac1K\sum_k\|r_k\|^2=\|\bar r\|^2+\frac1K\sum_k\|\delta_k\|^2$.
\end{proposition}

Geometrically, $\bar r$ moves the goal from $z_g$ to $z_g-\bar r$, so it can reorder candidates only when it is comparable to their distance from the goal.
At the starting models, \ThSharedMin--\ThSharedMax\% of the cost-error variance is a per-state constant.
Training can therefore lower the prediction error substantially while leaving the pick unchanged, as on Cube.

The pick is invariant to every positive affine map of the predicted costs.
We call the part of the cost error that survives the best such map the \emph{Decision-Relevant Residual} (DRR).

\begin{proposition}[The DRR bounds the latent regret]
\label{prop:drr}
Let $j=\arg\min_k\hat c_k$ and $k^\star=\arg\min_k c_k$. For $a>0$ and $b\in\mathbb R$, let $u_k=a\hat c_k+b-c_k$. Then
\begin{equation*}
c_j-c_{k^\star}\;\le\; 2\,\mathrm{DRR},\qquad
\mathrm{DRR}=\inf_{a>0,\,b}\,\max_k |u_k| .
\label{eq:drr}
\end{equation*}
\end{proposition}

\begin{corollary}[Margin]
\label{cor:margin}
If the two smallest true-future costs satisfy $c_{(2)}-c_{(1)}>2\,\mathrm{DRR}$, then $j=k^\star$ and the selection part of the regret is zero.
\end{corollary}

\noindent 
The DRR discards any constant shift and rescaling of the costs; the shared residual enters it only through the cross term of Proposition 1.
Proposition~2 bounds regret measured in the true-future latent cost,
$c_j-c_{k^\star}$, rather than task regret, $q_j-q_{k^\star}$.
Whether reducing this bound improves task outcomes depends on how
well the latent goal distance aligns with task cost, an alignment
that pretrained encoders do not guarantee.
We therefore test empirically whether DRR reduction predicts
selection gain (Sec.~5.2, Fig.~4).
For measurement we use a normalized root-mean-square counterpart of Proposition 2: the residual of the best fit $a\hat c + b$ to $c$ with $a \ge 0$ over the $K$ candidates of a state, divided by the standard deviation of $c$.
It equals $\sqrt{1 - \rho_s^2}$ when the within-state Pearson correlation $\rho_s$ of $\hat c$ and $c$ is positive, and 1 otherwise, where the best fit is constant.
States in which all candidates have the same true-future cost (at most 7 per pool) are conservatively assigned 1; excluding them changes none of the DRR correlations we report by more than 0.01. Proofs are in Supp.~B.

\textbf{Optimism among near-ties.}
The margin condition of \cref{cor:margin} fails when the best candidates are nearly tied, and the planner's final population concentrates exactly there.
If the residuals of $L$ near-tied candidates behave like independent Gaussian noise of scale $\sigma$, the argmin selects the most optimistic of $L$ draws, whose expected optimism grows like $\sigma\sqrt{2\log L}$, the optimizer's curse~\cite{smith2006optimizer}.
This is consistent with the saturation in \cref{fig:candcount}: more candidates mean more near-ties for the error to exploit.

\textbf{Implications.} The analysis makes two predictions, which we test on independent pools.
\textbf{(i)} A prediction gain improves selection to the extent that it lowers the DRR (\cref{sec:transfer}).
\textbf{(ii)} Shrinking the candidate-specific residual at selection time improves selection with the trained models as they are (\cref{sec:method}).
\section{Decision-Aligned Selection}
\label{sec:method}

\Cref{sec:theory} shows that the argmin acts on the candidate-specific part of the cost error.
Among near-tied candidates, it picks the most optimistic residual, and this optimism grows with the residual scale $\sigma$.
Continued training therefore improves selection only to the extent that it lowers this candidate-specific error.
We instead lower it at selection time, by combining checkpoints whose candidate-specific errors differ.
We call the resulting rule \emph{Decision-Aligned Selection} (DAS).

\subsection{Consensus Scoring}
\label{sec:consensus}

\textbf{Rule.}
Let $\theta_1,\dots,\theta_M$ be independently trained checkpoints, \eg, seeds of one training recipe or different recipes, with predicted costs $\hat c_m(k)$ for the candidates $k$ of a state.
Each checkpoint's costs are first standardized over the candidates:
\begin{equation*}
s_m(k)=\frac{\hat c_m(k)-\operatorname{mean}_{k'}\hat c_m(k')}{\operatorname{std}_{k'}\hat c_m(k')},
\label{eq:standardize}
\end{equation*}
with $s_m(k)=0$ if a checkpoint gives all candidates the same cost.
This positive affine map leaves each checkpoint's own choice unchanged (\cref{sec:theory}).
It also puts all checkpoints on a common scale, so that no checkpoint dominates the combination through its cost scale alone.
DAS then selects
\begin{equation*}
j_{\mathrm{DAS}}=\arg\min_k\;\Big[\,\underbrace{\tfrac1M{\textstyle\sum_m} s_m(k)}_{\text{consensus}}\;+\;\beta\,\underbrace{\operatorname{std}_m s_m(k)}_{\text{disagreement}}\Big].
\label{eq:das}
\end{equation*}

The consensus term is the core of the method.
The optional disagreement term adds pessimism toward candidates on which the checkpoints disagree~\cite{chua2018pets,jin2021pessimism}.
Before any test use, we fixed $M=3$ and $\beta=1$ on the development pools.
All main results use this frozen rule; \cref{sec:ablation} reports the rule without standardization and with other values of $\beta$ and $M$.

\textbf{Rationale.}
\label{sec:whyconsensus}
Let each checkpoint's residual after its best affine fit (\cref{prop:drr}) decompose as $u^{(m)}_k=u_k+\epsilon^{(m)}_k$, with $u_k$ shared by all checkpoints and $\epsilon^{(m)}_k$ independent across checkpoints with scale $\sigma_\epsilon$.
Averaging the fitted costs leaves the residual $u_k+\frac1M\sum_m\epsilon^{(m)}_k$, whose checkpoint-specific part has scale $\sigma_\epsilon/\sqrt M$.
Among near-tied candidates, this shrinks the optimism of the pick, while the shared part $u_k$ remains.
This argument is an idealization.
Real checkpoints share data, objective and initialization, so their errors are only partly independent.
The argument also ignores standardization and the disagreement term.
It still predicts that the gain grows with the share of the DRR that the checkpoints do not share, which we measure in \cref{sec:ablation}.
It also suggests that more diverse checkpoints, \eg, from different training recipes, gain more than seeds of one recipe.

\subsection{Integration into CEM}
\label{sec:cem}

DAS replaces only the scoring step of the planner (\cref{fig:overview}, panel~4).
At every CEM iteration, the $M$ checkpoints roll out the current population of $N{=}300$ samples.
The costs are standardized over that population (\cref{eq:standardize}), and the samples are ranked by \cref{eq:das}.
Elite selection, the update of the sampling distribution and plan execution are unchanged.
On fixed pools, the same rule is applied once to the $K{=}128$ frozen candidates.

DAS requires no training beyond the multi-seed or multi-recipe runs that produce its checkpoints.
Its cost is $M$ forward passes per sample.
Our implementation evaluates the checkpoints one after another, so planning time grows linearly with $M$.
On Wall (DINO-WM), a planning call takes 74\,s with $M=3$ and 24\,s with a single checkpoint.
\Cref{sec:closedloop} compares DAS with a single checkpoint given the same number of model evaluations ($MN=900$ samples), which separates the effect of combining models from that of searching harder with one.

\section{Experiments}
\label{sec:experiments}
\subsection{Experimental Setup}
\label{sec:setup}
\begin{table*}[t]
\centering
\small
\setlength{\tabcolsep}{4pt}
\begin{tabular}{lllccccc}
\toprule
Model & Task & Recipe & $\Delta$ latent err. & $\Delta$ pairwise acc.\ (pp) & $\Delta$ selected quality & \multicolumn{2}{c}{\% of gap} \\
\cmidrule(l){7-8}
 & & & & & & test & dev \\
\midrule
\multirow{12}{*}{LeWM~\cite{maes2026lewm}}
 & \multirow{3}{*}{PushT~\cite{florence2021ibc,chi2023diffusion}} & $B$ & $-8\%$ & $+1.4\,[-1.4, +4.2]$ & $-0.021\,[-0.063, +0.024]$ & \cellcolor{blue!6}$+6$ & $+8$ \\
 & & $O$ & $-9\%$ & $+2.1\,[-0.7, +5.0]$ & $-0.022\,[-0.058, +0.015]$ & \cellcolor{blue!7}$+7$ & $+15$ \\
 & & $P$ & $-17\%$ & $+5.9\,[+2.8, +8.9]$ & $-0.088\,[-0.130, -0.046]$ & \cellcolor{blue!26}$+26$ & $+29$ \\
\cmidrule(l){2-8}
 & \multirow{3}{*}{Two-Room~\cite{sobal2025pldm}} & $B$ & $-20\%$ & $+2.3\,[+0.2, +4.4]$ & $-0.21\,[-0.87, +0.41]$ & \cellcolor{blue!7}$+7$ & $+15$ \\
 & & $O$ & $-8\%$ & $+1.3\,[-0.6, +3.0]$ & $-0.54\,[-1.12, -0.09]$ & \cellcolor{blue!19}$+19$ & $+7$ \\
 & & $P$ & $-10\%$ & $+1.0\,[-1.2, +3.3]$ & $-0.24\,[-0.75, +0.26]$ & \cellcolor{blue!8}$+8$ & $-2$ \\
\cmidrule(l){2-8}
 & \multirow{3}{*}{Reacher~\cite{tassa2018dmc}} & $B$ & $-7\%$ & $+0.2\,[-0.9, +1.4]$ & $-2.2\,[-5.3, +0.8]$ & \cellcolor{blue!3}$+3$ & $-2$ \\
 & & $O$ & $-10\%$ & $+1.1\,[+0.0, +2.1]$ & $-4.2\,[-7.2, -1.3]$ & \cellcolor{blue!5}$+5$ & $-1$ \\
 & & $P$ & $-13\%$ & $+0.9\,[-0.8, +2.5]$ & $-4.8\,[-8.8, -1.0]$ & \cellcolor{blue!6}$+6$ & $-2$ \\
\cmidrule(l){2-8}
 & \multirow{3}{*}{Cube~\cite{park2025ogbench}} & $B$ & $-23\%$ & $-0.7\,[-1.7, +0.3]$ & $-0.13\,[-0.56, +0.30]$ & \cellcolor{blue!8}$+8$ & $-1$ \\
 & & $O$ & $-1\%$ & $-0.2\,[-0.7, +0.3]$ & $+0.27\,[-0.01, +0.58]$ & \cellcolor{red!17}$-17$ & $-4$ \\
 & & $P$ & $-1\%$ & $-0.3\,[-0.9, +0.3]$ & $+0.17\,[-0.19, +0.54]$ & \cellcolor{red!10}$-10$ & $+1$ \\
\midrule
\multirow{6}{*}{DINO-WM~\cite{zhou2024dinowm}}
 & \multirow{3}{*}{PushT~\cite{florence2021ibc,chi2023diffusion}} & $B$ & $-20\%$ & $+6.4\,[+5.1, +7.6]$ & $-0.144\,[-0.189, -0.104]$ & \cellcolor{blue!45}$+45$ & $+27$ \\
 & & $O$ & $-17\%$ & $+6.3\,[+5.0, +7.6]$ & $-0.144\,[-0.191, -0.102]$ & \cellcolor{blue!46}$+46$ & $+27$ \\
 & & $P$ & $-25\%$ & $+7.8\,[+6.0, +9.6]$ & $-0.180\,[-0.237, -0.127]$ & \cellcolor{blue!57}$+57$ & $+50$ \\
\cmidrule(l){2-8}
 & \multirow{3}{*}{Wall~\cite{zhou2024dinowm}} & $B$ & $-30\%$ & $+6.0\,[+4.0, +8.2]$ & $-0.25\,[-0.58, +0.05]$ & \cellcolor{blue!17}$+17$ & $+18$ \\
 & & $O$ & $-18\%$ & $+1.7\,[-0.5, +3.8]$ & $-0.05\,[-0.51, +0.42]$ & \cellcolor{blue!4}$+4$ & $+19$ \\
 & & $P$ & $-16\%$ & $+0.8\,[-2.2, +3.6]$ & $-0.20\,[-0.60, +0.17]$ & \cellcolor{blue!13}$+13$ & $+12$ \\
\bottomrule
\end{tabular}
\caption{\textbf{Training transfer on the test pools.} Each row is a starting model after 1{,}500 updates of one recipe (mean of 3 seeds). Quality is the task error at step 25 (lower is better; units: $d$ for PushT, px, mrad, grid cells, cm for Two-Room, Reacher, Wall, Cube). Brackets: 95\% hierarchical bootstrap intervals. \% of gap: share of the gap between the starting model's pick and the pick with true futures that training closes (shading: blue = gain, red = loss).}
\label{tab:test}
\vspace{-3mm}
\end{table*}

\textbf{Tasks and models.}
We use five simulated tasks: OGBench-Cube~\cite{park2025ogbench}, Two-Room~\cite{sobal2025pldm}, Reacher~\cite{tassa2018dmc}, PushT~\cite{florence2021ibc,chi2023diffusion} and Wall~\cite{zhou2024dinowm}.
The quality $q$ is the task error at step 25, \eg, the cube--goal distance on Cube and the normalized distance $d$ on PushT; success thresholds follow each benchmark (Supp.~A).
We evaluate the released checkpoints of two model families: LeWorldModel (LeWM)~\cite{maes2026lewm} on Cube, Two-Room, Reacher and PushT, and DINO-WM~\cite{zhou2024dinowm} on PushT and Wall (on Wall, scored with the visual term only, as in the official planning objective).
We omit DINO-WM on PointMaze~\cite{fu2020d4rl}, where selection has little headroom: the starting model’s pick is already near-optimal in most states (Supp.~A).
All models plan with their official CEM settings (300 samples, 30 elites, $H{=}25$).
Continued training (\cref{sec:prelim}) with all three recipes on the six releases yields 18 configurations (\cref{tab:test}).

\textbf{Pools and statistics.}
We split episodes once into train, development and test partitions; test pools are built like the development pools, with 192--314 states per task and 128 executed candidates per state.
All rules and hyperparameters were fixed on the development pools, and the analysis and nine hypotheses were frozen before any test pool was labeled (Supp.~C).
Each result is marked as pre-registered or exploratory; intervals are 95\% hierarchical bootstrap intervals over training seeds and episodes.
Closed-loop runs replan for 50 steps on 250 test scenes per task for LeWM and 100 for DINO-WM (Supp.~C), with the same scenes and planner seeds for every model and rule, so differences are paired by scene.

\textbf{Selectors and baselines.}
We compare DAS (\cref{eq:das}), over the seeds of one recipe or over several recipes, with the starting model and single checkpoints.
Three exploratory baselines use the same training compute: the best seed on the development pools, one model trained for 4{,}500 updates (three times the recipe budget), and the weight average of the three seeds~\cite{izmailov2018swa,wortsman2022soups}.
In closed loop, one checkpoint with 900 samples matches the planning compute of DAS (\cref{sec:closedloop}).

\subsection{Training Transfer}
\label{sec:transfer}

\textbf{Training transfers to decisions unevenly.}
Table~\ref{tab:test} reports all 18 configurations on the test pools: training lowers the latent prediction error in every one (by up to 30\%), but selection improves by very different amounts.
The clearest gains are on PushT in both model families.
As pre-registered (H1, H2; Table.~\ref{tab:hyp}), plain continued training ($P$) closes \TPushTPGap\% of the gap to selection with true futures with LeWM and \TPushTDinoPGap\% with DINO-WM, and raises the rate of near-optimal picks by 17 and 31 points.
Prediction failures, i.e., wrong picks that true futures would have avoided, fall from \TPushTPPredFailStart\% to \TPushTPPredFailTrained\% of picks with LeWM and from \TPushTDinoPPredFailStart\% to \TPushTDinoPPredFailTrained\% with DINO-WM.
On Two-Room, Reacher and Wall, training closes only 3--19\% of the gap, including \TTwoRoomBGap\% and \TWallBGap\% for the pre-registered configurations with $B$ (H3, H4).
On Cube, no recipe improves selection detectably; $O$ and $P$ move it in the wrong direction (by 17\% and 10\% of the gap).
Removing the executed candidates ($O$ vs.\ $B$) does not change selection detectably in any pair (H7; Supp.~C).
The development pools rank the configurations as the test pools do (Spearman $\rho=+0.69$).

\begin{figure}[h]
\centering
\includegraphics[width=\linewidth, height=0.6\linewidth]{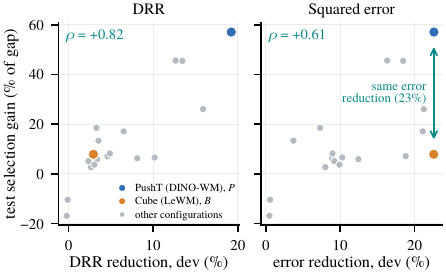}
\caption{\textbf{The DRR anticipates transfer.} Each point is one configuration: reduction on the development pools (horizontal) against selection gain on the test pools (vertical). Left: DRR (Prop.~\ref{prop:drr}), $\rho{=}{+}0.82$. Right: squared error, $\rho{=}{+}0.61$. PushT (DINO-WM) $P$ and Cube (LeWM) $B$ have equal squared-error reduction (23\%) but different DRR reduction and selection gain.}
\label{fig:predict}
\vspace{-4mm}
\end{figure}

\begin{table}[t]
\centering
\scriptsize
\setlength{\tabcolsep}{2.5pt}
\begin{tabular}{@{}lp{0.60\columnwidth}p{0.30\columnwidth}@{}}
\toprule
& Pre-registered prediction & Test result \\
\midrule
H1 & PushT, LeWM, $P$: selection improves & $-0.088\,d$ \\ 
H2 & PushT, DINO-WM, $P$: selection improves & $-0.180\,d$ \\ 
H3 & Two-Room, $B$: selection improves & $-0.21$\,px \\ 
H4 & Wall, $B$: selection improves & $-0.25$\,grid \\ 
H5 & Pairwise acc.\ predicts transfer better than error & $\rho{=}{+}0.73$ vs.\ ${+}0.61$ \\ 
H6 & Severe-miss rate unchanged & 3 of 18 decrease \\ 
H7 & $O$ and $B$ do not differ & 0 of 6 differ \\ 
H8 & DAS beats single in $\ge$ half of configs & 12/18, median ${+}1.6$ \\ 
H9 & DAS closed-loop success above single & mean ${+}0.5$\,pp (6/9) \\ 
\bottomrule
\end{tabular}
\caption{\textbf{Pre-registered hypotheses and test results}, frozen before any test pool was labeled (full wording and intervals in Supp.~C).} 
\label{tab:hyp}
\vspace{-4.5mm}
\end{table}
\textbf{The DRR anticipates which configurations transfer.}
Section~\ref{sec:theory} predicts that selection follows the reduction of the DRR, not of the squared error.
Measured on the development pools, the DRR reduction anticipates the test-pool selection gain at $\rho=+0.82$, against $+0.61$ for the squared-error reduction (Fig.~\ref{fig:predict}).
PushT with DINO-WM ($P$) and Cube ($B$) both lower the squared error by 23\%, but $P$ lowers the DRR by 19\% and closes 57\% of the gap, while $B$ lowers it by 3\% and closes 8\%.
The raw centred cost error barely tracks the gain ($\rho=+0.08$), as it also penalizes harmless rescaling. The pre-registered predictor, the change in pairwise accuracy, reaches $\rho=+0.73$ (H5; Table.~\ref{tab:hyp}).
Training improves selection to the extent that it reduces the part of the error that reorders candidates. 

\vspace{-8pt}
\subsection{Open-Loop Selection}
\label{sec:das}
\textbf{Consensus improves selection broadly.}
Applied to the three checkpoints of each recipe on the test pools, our frozen rule selects better than single checkpoints in 12 of 18 configurations, with a median gain of $+1.6$ points of the gap (H8; per-configuration results in Supp.~C). 
The gain comes from the consensus itself: averaging the standardized costs alone ($\beta=0$) does better still, and a third checkpoint adds to it (Sec.~\ref{sec:ablation}; Supp.~G). 
DAS also helps where training does not. With $B$, training closes only 8\% of the gap on Cube and \TWallBGap\% on Wall (Sec.~\ref{sec:transfer}), and DAS adds another $+11$ and $+10$ points there. Fig.~\ref{fig:teaser}E shows a Cube state on which every single checkpoint picks a failing plan while DAS does not. 
DAS changes only how the planner scores candidates and leaves training untouched, so it can be combined with each training recipe we test.

\begin{table}[h]
\centering
\footnotesize
\setlength{\tabcolsep}{2pt}
\begin{tabular}{@{}lcccc@{}}
\toprule
& \multicolumn{2}{c}{compute (vs.\ one ckpt.)} & \multicolumn{2}{c}{vs.\ single checkpoint} \\
\cmidrule(lr){2-3}\cmidrule(l){4-5}
Selector & training & inference & better in & median gain \\
\midrule
best seed, chosen on dev        & 3$\times$ & 1$\times$ & 10/18 & $+0.3$ \\
one model, 4,500 updates & 3$\times$ & 1$\times$ & 6/18  & $-3.1$ \\
weight averaging                & 3$\times$ & 1$\times$ & 13/18 & $+1.3$ \\
\rowcolor{blue!6} DAS (ours)           & 3$\times$ & 3$\times$ & 12/18 & $+1.6$ \\
\bottomrule
\end{tabular}
\caption{\textbf{Baselines on the test pools.} Gain over single-checkpoint selection, in points of the gap to selection with true futures; ``better in'': configurations with a positive gain. All selectors use the training compute of three checkpoints.} 
\label{tab:baselines}
\vspace{-3mm}
\end{table}

\textbf{Combining checkpoints outperforms one model.}
Table~\ref{tab:baselines} compares DAS with three exploratory baselines, all trained with the compute of three checkpoints.
Choosing the best seed on the development pools barely helps (better than a single checkpoint in 10 of 18 configurations, median $+0.3$ points).
DAS selects better than this seed in 12 of 18, so its gain comes from combining checkpoints, not from avoiding a bad one.
Training one model three times longer (4,500 updates) hurts (better in 6 of 18, median $-3.1$), and DAS selects better than it in 14 of 18 (median $+5.6$ points).
With a fixed set of training states, the extra updates overfit: for LeWM, the test latent error falls by 9.6\% after 4,500 updates against 10.7\% after 1,500, and pairwise accuracy rises by 1.0 against 1.2 points.
Combining the checkpoints helps in either form (Table~\ref{tab:baselines}), as expected if the residuals specific to each checkpoint cancel (Sec.~\ref{sec:consensus}).
But weight averaging needs checkpoints whose parameters can be averaged, whereas DAS needs only their predicted costs and can combine any checkpoints that score the same candidates.
Consensus cannot remove errors shared across checkpoints.
Our cross-recipe ablation suggests that consensus gains are larger
when a greater share of the residual error is independent across
checkpoints (Sec.~\ref{sec:ablation}).

\subsection{Closed-Loop Planning}
\label{sec:closedloop} 

In closed-loop, the planner replans from its own outcomes, so an early better choice can change every later state.
Table~\ref{tab:cl} reports the main comparisons; Supp.~D lists every model and selector.
\begin{table}[h]
\centering
\footnotesize
\setlength{\tabcolsep}{3pt}
\begin{tabular}{@{}lccc@{}}
\toprule
Configuration & Start & Success & $\Delta$ vs.\ start \\
\midrule
PushT (LeWM), $P$                        &  &      &  \\
\quad \textit{single checkpoint}                  &   88.0   & 89.6 & $+1.6\,[-0.7, +4.0]$ \\
\quad \textit{single, 3$\times$ samples}          &   88.0   & 89.3 & $+1.3\,[-1.6, +4.3]$ \\
\rowcolor{blue!6}\quad \textit{DAS (ours)}        &   88.0   & 92.0 & $+4.0\,[+1.2, +7.2]$ \\
\midrule
PushT (DINO-WM), $P$                     & 91.0 & 90.0 & $-1.0\,[-4.7, +2.3]$ \\
Two-Room (LeWM), $B$                     & 80.8 & 82.8 & $+2.0\,[-0.7, +4.8]$ \\
Wall (DINO-WM), $B$                      & 93.0 & 96.0 & $+3.0\,[+0.7, +6.0]$ \\
\bottomrule
\end{tabular}
\caption{\textbf{Closed-loop success (\%) on the test scenes.} On PushT (LeWM), only DAS beats the starting model detectably, by $+2.4\,[+0.4, +4.5]$ over single $P$ and $+2.7\,[+0.1, +5.3]$ over one checkpoint with its compute (3$\times$ samples). Paired by scene; 95\% bootstrap intervals.}
\label{tab:cl}
\vspace{-3mm}
\end{table}

\textbf{Continued training helps modestly in closed loop.}
We run the closed loop on 13 configurations: all recipes on PushT, Two-Room, Reacher and Wall, and $P$ on PushT (DINO-WM).
Training raises success over the starting model in 9 of them, most clearly on Wall with the pre-registered recipe $B$ ($+3.0$ [$+0.7$, $+6.0$] points).
Since success is near its ceiling on most tasks, we also measure the distance to the goal, which falls detectably in 4 configurations and rises detectably in none (Supp.~D).
On PushT (DINO-WM), success stays within noise, but $P$ brings the planner closer to the goal ($-0.12d$ in final distance), the closed-loop counterpart of H2.
As on the pools, training improves decisions by small and uneven amounts.

\begin{figure}[h]
  \centering
  \includegraphics[width=\columnwidth]{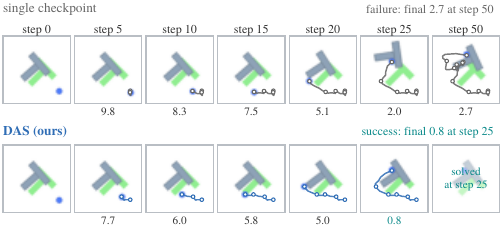}
  \caption{\textbf{Closed loop on PushT (LeWM), recipe $P$.}
Same scene and planner seed with a single checkpoint (top) and DAS (bottom). Lines: agent path; dots: replanning points; numbers: distance to the green goal (success below 1).}
  \label{fig:cl}
  \vspace{-3mm}
\end{figure}

\textbf{DAS turns open-loop gains into closed-loop success.}
On PushT (LeWM), DAS improves on single checkpoints for all three recipes. With $P$, it reaches 92.0\%, $+2.4\,[+0.4, +4.5]$ points over single checkpoints, which alone gain only 1.6 points over the starting model. Consensus thus carries the open-loop transfer into closed loop. 
The planners differ mainly near the goal, where candidates are nearly tied (Sec.~\ref{sec:theory}): by step 26, DAS has solved 88.0\% of the scenes, against 86.4\% for single $P$ checkpoints and 80.4\% for the starting model (Supp.~D). 
In Fig.~\ref{fig:cl}, the single checkpoint pushes the block past the goal after step 20, while DAS places it there at step 25. 
DAS also gives the best success on Reacher ($+2.4$ points over single checkpoints with $P$). 
Its gain follows how much decision-relevant error the seed checkpoints do not share (Sec.~\ref{sec:consensus}): it is largest on PushT (independent share 0.20) and smallest on Two-Room (0.10) and Wall (0.09; Supp.~D). 
Over the nine pre-registered LeWM comparisons, DAS beats single checkpoints by $\HNineMean$ points on average (H9), and with continued training it lies at or above the starting model in 11 of 12 configurations. 

\textbf{DAS beats a single checkpoint at equal compute.}
On PushT (LeWM), a single $P$ checkpoint given the same model evaluations as DAS (900 samples; logged before its results) stays at 89.3\%, whereas DAS reaches 92.0\% ($+2.7\,[+0.1, +5.3]$). 
Averaging the checkpoints' weights does not help either (89.6\%). 
More samples or merged weights still let the search exploit a single model's residual (Sec.~\ref{sec:theory}). Consensus averages the residuals of separate checkpoints, so they partly cancel.

\subsection{Component Ablations}
\label{sec:ablation}
\textbf{Varying the rule.}
On the test pools (Table.~\ref{tab:ablation}), consensus alone ($\beta=0$) beats a single checkpoint in 16 of 18 configurations (median $+2.2$ points).
Taking the median, averaging raw costs or summing ranks works almost as well (13 to 15 of 18).
A third checkpoint helps ($+1.6$ with two, $+2.2$ with three), as expected if averaging cancels checkpoint-specific residuals (\cref{sec:whyconsensus}).
The disagreement penalty lowers the median pool consensus
gain ($+1.6$ at $\beta=1$ vs.\ $+2.2$ at $\beta=0$), but yields
higher closed-loop success on PushT ($92.0\%$ vs.\ $90.8\%$;
Supp.~D). This suggests that penalizing disagreement helps
during replanning, although we evaluate this closed-loop
comparison only on PushT.
\begin{table}[h]
\centering
\footnotesize
\setlength{\tabcolsep}{4pt}
\begin{tabular}{@{}llcc@{}}
\toprule
 & Variant & better in & median gain \\
\midrule
\multicolumn{2}{@{}l}{consensus only (mean of z-scores, $\beta{=}0$, $M{=}3$)} & 16/18 & $+2.2$ \\
\midrule
\multirow{3}{*}{aggregation} & median of z-scores & 15/18 & $+2.3$ \\
 & mean of raw costs  & 15/18 & $+2.1$ \\
 & rank sum (Borda)   & 13/18 & $+1.2$ \\
\midrule
  & 0.5 & 14/18 & $+2.1$ \\
\rowcolor{blue!6}\cellcolor{white} & 1 (DAS) & 12/18 & $+1.6$ \\
\multirow{-3}{*}{penalty $\beta$} & 2 & 11/18 & $+0.7$ \\
\midrule
checkpoints $M$ & 2 & 14/18 & $+1.6$ \\
\bottomrule
\end{tabular}
\caption{\textbf{Rule ablations}}
\label{tab:ablation}
\vspace{-3mm}
\end{table}

\textbf{Measuring shared errors.}
Consensus removes only the error that checkpoints do not share (\cref{sec:whyconsensus}).
We measure this \emph{independent share} as one minus the correlation between two checkpoints' standardized residuals on the 13 truly best candidates of a state (top 10\%). 
It tracks the consensus gain across the 18 configurations ($\rho=0.70$ on the test pools and $0.64$ when measured on the development pools, $p\le0.004$; exploratory). 
Seeds of one recipe share most of their error (independent share 0--28\%), which limits seed ensembles; one checkpoint per recipe raises both the share and the gain in all six model--task pairs (median $+2.2$ points; Tab.~\ref{tab:diverse}), though no pair is significant alone. 

\begin{table}[h]
\centering
\footnotesize
\setlength{\tabcolsep}{3pt}
\begin{tabular}{@{}lc>{\columncolor{blue!6}}ccc@{}}
\toprule
 & \multicolumn{3}{c}{gain over single checkpoint} & indep.\ share \\
\cmidrule(lr){2-4}\cmidrule(l){5-5}
 & same & cross & cross $-$ same & same / cross \\
\midrule
\multicolumn{5}{@{}l}{\textit{LeWM}} \\
\quad PushT    & $+5.0$ & $+7.0$  & $+2.0\,[-1.5, +5.4]$   & 0.20 / 0.29 \\
\quad Two-Room & $+3.2$ & $+6.0$  & $+2.7\,[-5.8, +11.7]$  & 0.10 / 0.16 \\
\quad Reacher  & $+1.5$ & $+1.6$  & $+0.1\,[-0.9, +1.1]$   & 0.13 / 0.16 \\
\quad Cube     & $+2.7$ & $+5.1$  & $+2.4\,[-9.4, +15.2]$  & 0.07 / 0.13 \\
\midrule
\multicolumn{5}{@{}l}{\textit{DINO-WM}} \\
\quad PushT    & $+0.3$ & $+1.8$  & $+1.6\,[-3.2, +6.2]$   & 0.01 / 0.16 \\
\quad Wall     & $+7.5$ & $+14.7$ & $+7.2\,[-1.1, +17.1]$  & 0.09 / 0.32 \\
\bottomrule
\end{tabular}
\caption{\textbf{Cross-recipe consensus} (test pools, $\beta{=}0$). Gain over single-checkpoint selection, in points of the gap. \emph{same}: three seeds of one recipe, mean over recipes; \emph{cross}: one checkpoint per recipe, mean over seed indices. Brackets: 95\% bootstrap intervals.}
\label{tab:diverse}
\vspace{-6mm}
\end{table}

\vspace{-5pt}
\section{Conclusion}
\label{sec:conclusion}
Planning gains follow the DRR more closely than prediction error, and training improves decisions where it reduces the DRR.
DAS reduces it at selection time by combining checkpoints, and beats spending the same compute on one model.
We recommend reporting the DRR on fixed pools of executed candidates.
DAS costs $M$ times more inference and cannot repair errors in the objective, and our closed-loop tasks are short-horizon with success near ceiling.
Supp.~I discusses further limitations.
\section*{LLM Usage Statement}
We used Claude Opus 5.5 (Anthropic) to refine wording, shorten and restructure sentences and paragraphs, and format tables in \LaTeX.
Claude also reviewed and debugged parts of the experiment and analysis code.
All experiments were run by the authors, and every reported number comes from the authors' experiment logs.
The authors reviewed all AI-assisted content and take full responsibility for the paper.
{
    \small
    \bibliographystyle{ieeenat_fullname}
    \bibliography{main}
}
\clearpage
\end{document}